\def\year{2022}\relax
\pdfoutput=1
\documentclass[letterpaper]{article} % DO NOT CHANGE THIS
\usepackage{aaai22}  % DO NOT CHANGE THIS
\usepackage{times}  % DO NOT CHANGE THIS
\usepackage{helvet}  % DO NOT CHANGE THIS
\usepackage{courier}  % DO NOT CHANGE THIS
\usepackage[hyphens]{url}  % DO NOT CHANGE THIS
\usepackage{graphicx} % DO NOT CHANGE THIS
\usepackage{amsmath}
\usepackage{amsfonts}
\usepackage{booktabs}
\usepackage{natbib}  % DO NOT CHANGE THIS AND DO NOT ADD ANY OPTIONS TO IT
\usepackage{caption} % DO NOT CHANGE THIS AND DO NOT ADD ANY OPTIONS TO IT
\DeclareCaptionStyle{ruled}{labelfont=normalfont,labelsep=colon,strut=off} % DO NOT CHANGE THIS
\usepackage{algorithm}
\usepackage{algorithmic}

\usepackage{newfloat}
\usepackage{listings}
\floatstyle{ruled}
\newfloat{listing}{tb}{lst}{}
\floatname{listing}{Listing}
\title{Exploring Entity Interactions for Few-Shot Relation Learning (Student Abstract)}
\author{
    Yi Liang\textsuperscript{\rm 1}, 
    Shuai Zhao\textsuperscript{\rm 1},
    Bo Cheng\textsuperscript{\rm 1}, 
    Yuwei Yin\textsuperscript{\rm 2},
    Hao Yang\textsuperscript{\rm 3}
}
\affiliations{
    \textsuperscript{\rm 1} State Key Laboratory of networking and switching technology, \\Beijing University of Posts and Telecommunications, Beijing, China \\
    \textsuperscript{\rm 2} 2012 Labs, Huawei Technologies, CO., LTD, Shenzhen, China \\
    \textsuperscript{\rm 3} 2012 Labs, Huawei Technologies, CO., LTD, Beijing, China
    \{liangyi, zhaoshuaiby, chengbo\}@bupt.edu.cn, \\
    \{yinyuwei, yanghao30\}@huawei.com
}

\usepackage{bibentry}
\begin{document}

\maketitle

\begin{abstract}
Few-shot relation learning refers to infer facts for relations with a limited number of observed triples. Existing metric-learning methods for this problem mostly neglect entity interactions within and between triples. In this paper, we explore this kind of fine-grained semantic meanings and propose our model TransAM. Specifically, we serialize reference entities and query entities into sequence and apply transformer structure with local-global attention to capture both intra- and inter-triple entity interactions. Experiments on two public benchmark datasets NELL-One and Wiki-One with 1-shot setting prove the effectiveness of TransAM.
\end{abstract}

\section{Introduction}
Knowledge graphs (KGs) store facts in the form of triples $(h, r, t)$ reflecting the relation $r$ between head $h$ and tail entity $t$. Few-shot relation learning or few-shot knowledge graph completion (FSKGC), which aims to complete the relations with only a limited number of observed triples (few-shot references), has attracted research attention in recent years. However, existing FSKGC models mostly fail to model entity level interactions within and between triples as they perform matching over entity pair representations. 

In this paper, we argue that entity interactions can provide valuable fine-grained semantic meanings and propose TransAM, \textbf{T}ransformer\cite{Vaswani2017AttentionIA} \textbf{A}ppending \textbf{M}atcher with local-global attention for FSKGC. In particular, we serialize references of few-shot relation $r$ and query pair following with special token \texttt{[CLS]} into entity sequence and feed to transformer. To capture intra-triple level entity interactions, we construct a block attention mask matrix to constrain that every entity only attends to triple it involves (local attention). Besides, we introduce rotary operation to encode head or tail roles for each entity to model relation patterns, \emph{i.e.}, symmetry/anti-symmetry or inversion. Global attention focuses more on modelling inter-triple entity interactions. We devise a separated triple position encoding for the self-attention module to preserve the triple structure and decouple entity representations and triple position embedding as we concern they are two kinds of heterogeneous features. Finally, the representation of \texttt{[CLS]} is used to predict the plausibility of entity sequence, which indicates whether query fact $(h_q, r, t_q)$ holds.

\section{Proposed Method}
\paragraph{Task Definition} Knowledge graph $\mathcal{G}$ is formulated as $\mathcal{G} := \{\mathcal{E}, \mathcal{R}, \mathcal{T}\}$. $\mathcal{E}$ is entity set, $\mathcal{R}$ is a set of relations and $\mathcal{T}$ is the triple set. Background graph $\mathcal{BG}$ is a set of known triples. Giving a new relation $r$ with $K$ entity pairs as support set $\mathcal{S}_r=\{(h_i, t_i)\vert (h_i, r, t_i) \in \mathcal{G}\}$ and a query entity pair $(h_q, t_q)$, the goal of FSKGC is to estimate whether $(h_q, r, t_q)$ is a missing fact. \emph{In this work, we solve this problem by serializing support set and query pair to an entity sequence $\textbf{s}_q:=[\texttt{[CLS]}, h_1, t_1, \dots, h_K, t_K, h_{q}, t_q]$ and train the model to compute the probability of whether $\textbf{s}_q$ holds.}

\paragraph{Entity Encoder} Following previous work\cite{Zhang2020FewShotKG}, we enhance entity representations with a heterogeneous graph encoder. For each entity $e$ from input sequence $\textbf{s}_q$, after extracting neighbors $\mathcal{N}_e=\{(r_i, t_i)\}$ starting with $e$ from $\mathcal{BG}$, we calculate the contribution $\alpha_i = \frac{\exp(\textbf{u}^{T}(\text{ReLU} (\textbf{W}_1(\textbf{v}_{r_i}\Vert \textbf{v}_{t_i}))) + b_1)}{\sum_{j=1}^{\vert \mathcal{N}_e \vert} {\exp(\textbf{u}^{T}(\text{ReLU} (\textbf{W}_1(\textbf{v}_{r_j}\Vert \textbf{v}_{t_j}))) + b_1)}}$. Then we aggregate neighbors information following $\textbf{h}_{e} = \sum_{i=1}^{\vert \mathcal{N}_e \vert} {\alpha_i \textbf{v}_{t_i}}$. Further, a fusion mechanism is applied to preserve origin embedding $\textbf{v}_e$ and obtain the final representation $\textbf{x}_e= \sigma (\textbf{W}_2\textbf{v}_e + \textbf{W}_3\textbf{h}_e)$. Here, $\Vert$ denotes concatenate operation. $\sigma$ is non-linear function, and we use $\tanh$. $\textbf{u} \in \mathbb{R}^{d_e \times 1}$, $b_1 \in \mathbb{R}$, $\textbf{W}_1 \in \mathbb{R}^{d_e \times 2d_e}$, $\textbf{W}_2 \in \mathbb{R}^{d_e \times d_e}$ and $\textbf{W}_3 \in \mathbb{R}^{d_e \times d_e}$ are learnable parameters. We pack the encoder output of each entity of $\textbf{s}_q$ into matrix $\textbf{X}_q$. $d_e$ is the embedding dimension.

\begin{figure}[t]
    \centering
    \includegraphics[scale=0.8]{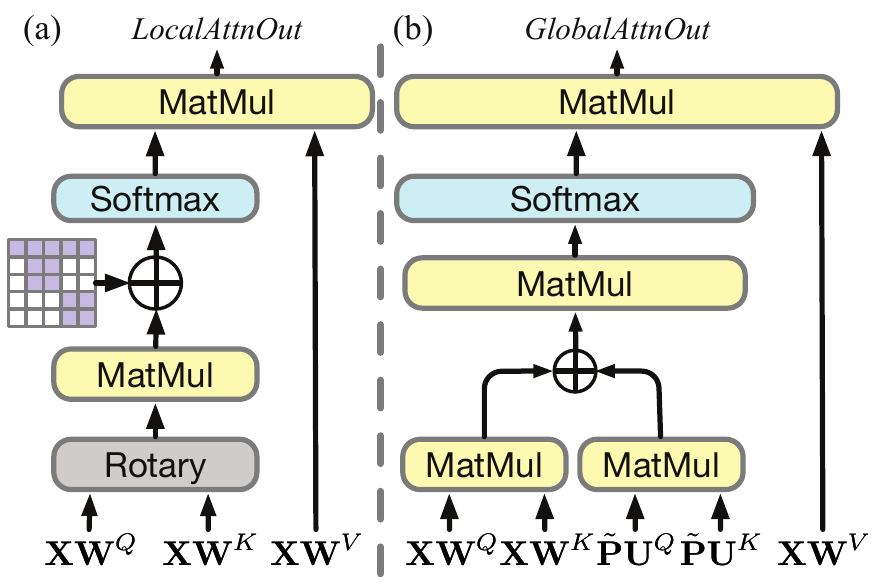}
    \caption{(a) Details of local attention; (b) Details of global attention. We sum \emph{LocalAttnOut} and \emph{GlobalAttnOut} and feed it to the rest structure of transformer.}
    \label{fig:archi}
\end{figure}

\paragraph{Transformer Matching Processor} Transformer Matching Processor contains a stack of $L$ identical blocks which mainly includes multi-head local-global attention module, position-wise feed-forward network (FFN) and layer normalization (LayerNorm). For $\textbf{X}_q$, the multi-head local-global attention (MHA) can be denoted as $\text{MHA}(\textbf{X}_q) = \mathop{\Vert}_{i=1}^{H}(Attn_{\text{global}}(\textbf{X}_q)+Attn_{\text{local}}(\textbf{X}_q))\textbf{W}^O$, where $Attn_{\text{global}}(\cdot)$ and $Attn_{\text{local}}(\cdot)$ indicate global attention and local attention function we describe later. $H$ is the number of attention heads. $\textbf{W}^O \in \mathbb{R}^{Hd_e\times Hd_{e}}$ is learnable projection matrix. Fig.~\ref{fig:archi} illustrates the architectures of local and global attention. Please refer to \cite{Vaswani2017AttentionIA} for more details about FFN and LayerNorm.
% $\textbf{W}^Q \in \mathbb{R}^{Hd_{e}\times d_e}$, $\textbf{W}^K \in \mathbb{R}^{Hd_{e}\times d_e}$, $\textbf{W}^V \in \mathbb{R}^{Hd_{e}\times d_e}$ are trainable projection matrices. 
\paragraph{Local Attention} Firstly, we construct a block attention mask matrix $\textbf{M}_{\text{local}}$ to enable self-attention module capturing intra-triple entity interaction features. We further utilize rotary operation $f_{R}^{\{Q, K\}}(e, m)=(\textbf{W}^{\{Q, K\}}\textbf{x}_{e})e^{im\theta}$ for each entity $e$ to endow entities with role information (head or tail) within triple, where $m$ is 1 and 2 for head and tail, respectively. Finally, the local attention output is computed as $\textbf{H}_{\text{local}} = \text{Softmax}(\frac{f_R^{Q}(\textbf{X}_q)\cdot f_R^{K}(\textbf{X}_q)^T}{\sqrt{d}}+\textbf{M}_{\text{local}})\textbf{X}_q\textbf{W}^V$. $\textbf{M}_{\text{local}}$ is obtained follows:
\begin{equation}
    \begin{split}
        m_{i, j} &= \begin{cases}
             & i=0 \text{ or } i=j \text{ or }\\
             0, & j=i+1, i \in \{1, \dots, 2K+1\} \text{ or }\\
             & j=i-1, i \in \{2, \dots, 2K+2\}; \\
            -\inf, & \text{else}
        \end{cases}
    \end{split}
\end{equation}

\paragraph{Global Attention} We first assign position $i$ for both entity $h_i$ and $t_i$ of triple $(h_i, t_i)$, while the position for \texttt{[CLS]} and query pair $(h_q, t_q)$ is set to 0 and $K+1$, respectively. A $d_e$ dimensional learnable embedding is applied to obtain representation of each position then we pack them into a matrix $\tilde{\textbf{P}}$. Finally, we compute the global stage hidden state following $\textbf{H}_{\text{global}} = \text{Softmax}(\textbf{A}_g)\textbf{X}_q\textbf{W}^V$, where $\textbf{A}_g = \frac{(\textbf{X}_q\textbf{W}^Q)(\textbf{X}_q\textbf{W}^K)^T + (\tilde{\textbf{P}}\textbf{U}^Q) (\tilde{\textbf{P}}\textbf{U}^K)^T}{\sqrt{2d_e}}$. 

\paragraph{Prediction and Optimization} Assuming the last hidden state of \texttt{[CLS]} is $\textbf{z}^L_{CLS}$, we obtain final score $\bar{\textbf{s}}_q$ of $(h_q, t_q)$ follow $\bar{\textbf{s}}_q = \text{Softmax}(\textbf{U}^T_2(\textbf{W}_4 \textbf{z}^L_{\text{CLS}}))$, in which $\textbf{W}_4 \in \mathbb{R}^{d_e\times Hd_e}$, $\textbf{U}^T_2 \in \mathbb{R}^{d_e\times 2}$. $\bar{\textbf{s}}_q \in \mathbb{R}^2$ is a 2-dimensional real vector while $s_q^{(1)}$ indicates the probability of whether query $(h_q, r, t_q)$ holds. We calculate loss $\mathcal{L} = -\sum_{q\in \mathbb{S}^+ \cup \mathbb{S}^-} ((1-y_q) \log(s_q^{(0)}) + y_q\log (s_q^{(1)}))$, where $\mathbb{S}^+$ and $\mathbb{S}^-$ indicate positive and negative query sequence respectively. $y_q \in \{0, 1\}$ is the sequence label (negative or positive).

\section{Experiments}
\paragraph{Datasets and Baselines} We use two public datasets NELL-One and Wiki-One proposed by GMatching\cite{Xiong2018OneShotRL}. We select Mean Reciprocal Rank (MRR), Hits@1 (H1) and Hits@10 (H10) as evaluation metrics. The few-shot size $K$ is set to 1, \emph{i.e.}, one-shot. We compare TransAM with three baselines most relevant to our research, GMatching\cite{Xiong2018OneShotRL}, FSRL\cite{Zhang2020FewShotKG} and FAAN\cite{Sheng2020AdaptiveAN}. Results of FSRL and FAAN are obtained using their official implementations with recommended hyperparameters reported in their papers.

\paragraph{Implementation Details} We choose ComplEx pre-trained embedding vectors to initialize all entities and relations. Embedding dimensionality is 100 for NELL-One and 50 for Wiki-One. We set the number of Transformer layers to 3 and 4, and the number of heads to 4 and 8 for NELL-One and Wiki-One, respectively. We warm up the model at the first 10k steps by linearly increasing the learning rate to $5e^{-5}$ for NELL-One and $6e^{-5}$ for Wiki-One, then decreasing it linearly to 0 until the last step.

\paragraph{Results and Conclusions} From table~\ref{tab:Result}, we observe that TransAM outperforms all three baselines in MRR and Hits@1 on both two datasets. Giving only one instance, matching models are expected to capture accurate semantic meanings for prediction. These results reveal that TransAM is appropriate for addressing few-shot issue in knowledge graph completion. Our future work may consider introducing sophisticated structural bias to the transformer for complex few-shot relations.
\begin{table}
    \centering
    \scalebox{0.9}{
        % \begin{tabular}{l|ccc|ccc}
        \begin{tabular}{l|p{2em}p{2em}p{2em}|p{2em}p{2em}p{2em}}
            \toprule
            & \multicolumn{3}{c}{NELL-One} & \multicolumn{3}{c}{Wiki-One} \\
            \cmidrule{2-4} \cmidrule{5-7}
            Model & \multicolumn{1}{c}{H1} & \multicolumn{1}{c}{H10} & \multicolumn{1}{c}{MRR} & \multicolumn{1}{c}{H1} & \multicolumn{1}{c}{H10} & \multicolumn{1}{c}{MRR} \\
            \midrule
            GMatching$^\dagger$ & .119 & .313 & .185 & .120 & .336 & .200 \\
            FSRL & .114 & .294 & .170 & .091 & .267 & .149 \\
            FAAN & .116 & .316 & .187 & .146 & \textbf{.362} & .215 \\
            \midrule
            TransAM & \textbf{.152} & \textbf{.360} & \textbf{.225} & \textbf{.184} & .358 & \textbf{.242} \\
            \midrule
        \end{tabular}
    }
    \caption{Experiment Results on NELL-One and Wiki-One. Best results are boldfaced. $^\dagger$ indicates results are taken from the origin paper.}
    \label{tab:Result}
\end{table}

\section{Acknowledgments}
This work is supported by Beijing Nova Program of Science and Technology (Grant No. Z191100001119031), Guangxi Key Laboratory of Cryptography and Information Security (No.GCIS202111). Bo Cheng is the corresponding author. 

\bibliography{custom.bib}

\end{document}